\documentclass{article}

\usepackage{arxiv}
\usepackage[utf8]{inputenc} 
\usepackage[T1]{fontenc}    
\usepackage{hyperref}       
\usepackage{url}            
\usepackage{booktabs}       
\usepackage{amsfonts}       
\usepackage{nicefrac}       
\usepackage{microtype}      
\usepackage{lipsum}

\usepackage{rotating}
\usepackage{multirow}
\usepackage{graphicx}
\usepackage{multicol}
\usepackage{comment}
\usepackage{amsmath}
\usepackage{booktabs}
\usepackage{makecell}
\usepackage{tabularx}
\usepackage[title,toc,titletoc,page]{appendix}
\usepackage[caption=false]{subfig}
\usepackage{pgfplots}
\usepackage{algorithm}
\usepackage{algpseudocode}

\algnewcommand{\LeftComment}[1]{\State \(\triangleright\) #1}

\algtext*{EndWhile}
\algtext*{EndIf}
\algtext*{EndFor}
\algtext*{EndFunction}

\newcommand{\eg}{{\textit{e.g.}}}
\newcommand{\ie}{{\textit{i.e.}}}
\newcommand{\etal}{{\textit{et al.}}}

\title{Neuroevolutionary Transfer Learning of Deep Recurrent Neural Networks through Network-Aware Adaptation}

\author{
    AbdElRahman ElSaid\thanks{Rochester Institute of Technology, Rochester, New York }  \\
    \texttt{aelsaid@mail.rit.edu}   \\
\And
    Joshua Karns\footnotemark[1]    \\
    \texttt{josh@mail.rit.edu}   \\
\And
    Zimeng Lyu\footnotemark[1]  \\
    \texttt{zimenglyu@mail.rit.edu}  \\
\And
    Daniel Krutz\footnotemark[1]    \\
    \texttt{dxkvse@rit.edu} \\
\And
    Alexander Ororbia II\footnotemark[1]    \\
    \texttt{ago@cs.rit.edu}  \\
\And
    Travis Desell\footnotemark[1]      \\
    \texttt{tjdvse@rit.edu} \\
}

\begin{document}
\maketitle

\begin{abstract}
    Transfer learning entails taking an artificial neural network (ANN) that is trained on a source dataset and adapting it to a new target dataset. While this has been shown to be quite powerful, its use has generally been restricted by architectural constraints. Previously, in order to reuse and adapt an ANN's internal weights and structure, the underlying topology of the ANN being transferred across tasks must remain mostly the same while a new output layer is attached, discarding the old output layer's weights.  This work introduces \emph{network-aware adaptive structure transfer learning} (N-ASTL), an advancement over prior efforts to remove this restriction. N-ASTL utilizes statistical information related to the source network's topology and weight distribution in order to inform how new input and output neurons are to be integrated into the existing structure. Results show improvements over prior state-of-the-art, including the ability to transfer in challenging real-world datasets not previously possible and improved generalization over RNNs trained without transfer.



\end{abstract}


\section{Introduction}
\label{sec:introduction}

As predictive analytics becomes increasingly more commonplace, there is a widespread and growing number of real-world systems that stand to benefit from an improved ability to forecast and predict data. In many fields, such as power systems, aviation, transportation, and manufacturing, developing engines for predictive analytics requires time series forecasting algorithms capable of adapting to noisy, constantly changing sensors as well as major system modifications or upgrades. Mechanical systems, such as self-driving cars, robotics, aircraft, and unmanned aerial systems (UAS) in particular would be significantly improved from the ability to more accurately predict potential equipment and systems failure, which is critically important for cost and safety reasons. Developing predictive models for these systems poses a particularly challenging problem given that these systems experience rapid changes in terms of their operation and sensor capabilities.

Transfer learning potentially offers a solution. It has already proven to be a powerful tool to improve the optimization of deep artificial neural networks (ANNs), allowing them to re-use and fine-tune knowledge gained from training on large, prior datasets. However, transfer learning has mostly been limited to problems which require minimal to no topological changes, \ie, changing the number of neurons and/or their synaptic connectivity, so that the previously trained weights and structure can be more easily fine tuned to the new target dataset. This is common in ANN specialization, often referred to as ``pre-training''. Gupta~\etal~ specialized a recurrent neural network (RNN) trained to predict $20$ different phenotypes from clinical data, by retraining it to predict previously unseen phenotypes with limited data~\cite{gupta2018transfer}. Zhang~\etal~applied the same principle when predicting the remaining useful life (RUL) of various systems when data was scarce~\cite{zhang2018transfer}. Other examples include ~\cite{yoon2017efficient, zarrella2016mitre, zhang2018transfer, mrkvsic2015multi}.

Approaches involving some structural network modification include work by Mun~\etal, which removed an ANN's output layer, replacing it with two additional hidden layers and a new output layer~\cite{mun2017deep}. Partial knowledge transfer has also been done through the use of pre-trained word and phrase embeddings~\cite{mikolov2013distributed,devlin2018bert}. Hinton~\etal~proposed the concept of ``knowledge distillation'', where an ensemble of teacher models are ``compressed'' to a single pupil model~\cite{hinton2015distilling}. Tang~\etal~also conducted the converse of this experiment -- they trained a complex pupil model using a simpler teacher model~\cite{tang2016recurrent}. Their findings demonstrated that knowledge gathered by a simple teacher model can be transferred to a more complex pupil model yielding greater generalization capability. Deo~\etal~also concatenated mid-level pre-trained features from two datasets as input to a target feedforward network~\cite{deo2017stacked}. 

Neuroevolutionary approaches have mostly focused on using indirect encodings for transfer learning, such as Taylor~\etal~utilizing NeuroEvolution of Augmenting Topologies (NEAT~\cite{stanley2002evolving}) to evolve task mappings to translate trained neural network policies between a source and target~\cite{taylor2007transfer}. Yang~\etal~took this concept further by designing networks for cross-domain, cross-application, and cross-lingual transfer settings~\cite{yang2017transfer}. Vierbancsics~\etal~used a different approach where an indirect encoding was evolved that could be applied to ANN tasks that required different neural structures~\cite{verbancsics2010evolving}.

To date, transfer learning research has almost exclusively focused on classification tasks, with most work focusing on feedforward and convolutional neural networks (CNNs), and has mostly avoided any major architectural changes to the ANNs being transferred. While some studies have utilized recurrent neural networks (RNNs), they have not attempted to develop RNN models for time series data forecasting, a very challenging regression problem (see Section~\ref{sec:data} for example prediction problems and why traditional statistical auto-regressive forecasting methods like ARIMA are insufficient). This work advances transfer learning to this area, as these capabilities will play a crucial role in developing predictive engines for previously described systems.

Prior work introduced \emph{adaptive structure transfer learning} (ASTL), where input and output nodes were added and removed to adapt the structure of an RNN evolved on a source data set to a target data set~\cite{elsaid2019evostar-transfer}. However, this work did not take into account the overall structure and weight distribution of the transferred network, only adding minimal connections for the newly added input and output nodes. This work proposes a significantly improved approach for transferring RNN knowledge across tasks through \emph{network-aware adaptive transfer learning} (N-ASTL). N-ASTL overcomes the limitations of ASTL by utilizing statistical information related to the source RNN's topology and weight distribution to inform the adaptation process. N-ASTL shows significant improvements over previously reported results, including successful transfer learning on the challenging datasets where ASTL previously failed. 
 
\section{Evolutionary eXploration of Augmenting Memory Models}

This work utilizes the Evolutionary eXploration of Augmenting Memory Models (EXAMM) algorithm~\cite{ororbia2019examm} to drive the neuroevolution process. EXAMM evolves progressively larger RNNs through a series of mutation and crossover (reproduction) operations. Mutations can be edge-based: \emph{split edge}, \emph{add edge}, \emph{enable edge}, \emph{add recurrent edge}, and \emph{disable edge} operations, or work as higher-level node-based mutations: \emph{disable node}, \emph{enable node}, \emph{add node}, \emph{split node} and \emph{merge node}. The type of node to be added is selected uniformly at random from a suite of simple neurons and complex memory cells: $\Delta$-RNN units~\cite{ororbia2017diff}, gated recurrent units (GRUs)~\cite{chung2014empirical}, long short-term memory cells (LSTMs)~\cite{hochreiter1997long}, minimal gated units (MGUs)~\cite{zhou2016minimal}, and update-gate RNN cells (UGRNNs)~\cite{collins2016capacity}. This allows EXAMM to select for the best performing recurrent memory units. EXAMM also allows for \emph{deep recurrent connections} which enables the RNN to directly use information beyond the previous time step. These deep recurrent connections have proven to offer significant improvements in model generalization, even yielding models that outperform state-of-the-art gated architectures~\cite{desell2019evostar-deeprecurrent}.  To the authors' knowledge, these capabilities are not available in other neuroevolution frameworks capable of evolving RNNs, which is the primary reason EXAMM was selected to serve as the foundation for this work. Due to space limitations we refer the reader to Ororbia \etal~\cite{ororbia2019examm} for more details on the EXAMM algorithm. The N-ASTL and ASTL implementations have been made freely available and incorporated into the EXAMM github repository\footnote{https://github.com/travisdesell/exact}.


To speed up the neuro-evolution process, EXAMM utilizes an asynchronous, distributed computing strategy that incorporates the concept of islands to promote speciation. This mechanism encourages both exploration and exploitation of massive search spaces. A master process maintains the populations for each island and generates new RNN candidate models from the islands in a round-robin manner. Workers receive candidate models and locally train them with back-propagation through time (BPTT), making EXAMM a memetic algorithm. When a worker completes the training of an RNN, that RNN is inserted back into the island that it originated from. Then, if the number of RNNs in an island exceeds the island's maximum population size, the RNN with the worst fitness score, i.e., validation set mean squared error (MSE), is deleted.

This asynchronous approach is particularly important given that the generated RNNs will have different topologies, with each candidate model requiring a different amount of time to train. This strategy allows the workers to complete the training of the generated RNNs at whatever speed they are capable of, yielding an algorithm that is naturally load-balanced.  Unlike synchronous parallel evolutionary strategies, EXAMM easily scales up to any number of available processors, allowing population sizes that are independent of processor availability. The EXAMM codebase has a multi-threaded implementation for multi-core CPUs as well as an MPI~\cite{mpif94mpi} implementation that allows EXAMM to readily leverage high performance computing resources.

To initialize the island populations, EXAMM ``seeds'' each island population with the minimal network topology possible for the given inputs and outputs, i.e., a topology with no hidden nodes where each input node has a single feed forward connection to each output node. Each island population utilizes this \emph{minimal genome} as a seed network as the first RNN in its population, which is ultimately sent to the first worker requesting an RNN to be trained. Subsequent requests for work from that island create new RNN candidates from mutations of the seed network until the population is full. When an island population is full, EXAMM will start generating new RNNs from that island utilizing both mutation and intra-island crossover (both parents are selected within that same island). When all island populations are full, EXAMM will then generate additional, new RNNs from an inter-island crossover process. This crossover selects the first parent from the island that an RNN is being generated from and matches it with the most fit RNN from another, randomly selected island to serve as the second parent.
\begin{algorithm}[ht]
    \begin{algorithmic}
        \Function{RemoveUnused}{SeedNetwork sn, Param[] targetOutputs, Param[] targetInputs}
            \LeftComment{Remove unused input and output nodes}
            \ForAll{InputNode i in sn.inputNodes}
                \If{i.param not in targetInputs}
                    \State sn.removeNode(i)
                \EndIf
            \EndFor

            \ForAll{OutputNode o in sn.outputNodes}
                \If{o.param not in targetOutputs}
                    \State sn.removeNode(o)
                \EndIf
            \EndFor

            \LeftComment{Mark reachability of edges and hidden nodes}
            \State sn.markForwardReachability()
            \State sn.markBackwardReachability()

            \ForAll{Node n in sn.hiddenNodes}
                \If{!n.forwardReachable() or !n.backwardReachable()}
                    \State n.setDisabled()
                \EndIf
            \EndFor

            \ForAll{Edge e in sn.edges}
                \If{!e.forwardReachable() or !e.backwardReachable()}
                    \State e.setDisabled()
                \EndIf
            \EndFor
        \EndFunction
    \end{algorithmic}
    \caption{\label{alg:remove_unused} Removal of Unused Nodes and Edges}
\end{algorithm}

\section{Network-Aware Adaptive Structure Transfer Learning}
\label{sec:examm_transfer_learning}

\subsection{Adaptive Structure Transfer Learning}
\label{sec:adaptive_transfer}
Prior work on \emph{adaptive structure transfer learning} (ASTL) proposed a simple scheme for transfer learning in EXAMM by hijacking the island seeding process to ultimately modify a previously trained RNN instead of the minimal genome~\cite{elsaid2019evostar-transfer}. The previously trained RNN is itself adapted to a new dataset through the following steps: {\it i)} remove unused outputs, {\it ii)} connect new outputs to all inputs, {\it iii)} remove unused inputs, {\it iv)} connect new inputs to all outputs, and {\it v)} disable \emph{vestigial} hidden nodes and edges.  

The disabling of vestigial structures is crucial since removing the unused inputs and outputs can potentially disconnect parts of the RNN's topology, which, if retained, would yield wasted computation. To safeguard against this, all edges and nodes are flagged for forward reachability, \ie, there is a path to the edge or node from an enabled input node, and backward reachability, \ie, there is a path from the node or edge to any output. Nodes and edges which are not forward and backward reachable are labeled as vestigial and disabled. They can, however, later be reconnected and enabled via EXAMM's mutation/crossover operations, essentially boostrapping the learning process by enabling easy reuse of previously learned neural circuits. This process is formalized in Algorithms~\ref{alg:remove_unused} and~\ref{alg:original_transfer}.

\begin{algorithm}[ht]
    \begin{algorithmic}
        \Function{ASTL}{SeedNetwork sn, Param[] targetOutputs, Param[] targetInputs}
            \State \Call{RemoveUnused}{sn, targetOutputs, targetInputs}
        
            \LeftComment{Add new input and output nodes}
            \ForAll{Param ti in targetInputs}
                \If{!sn.hasInputForParam(ti)}
                    \State sn.addInputNode(new InputNode(ti))
                \EndIf
            \EndFor

            \ForAll{Param to in targetOutputs}
                \If{!sn.hasOutputForParam(to)}
                    \State sn.addOutputNode(new OutputNode(to))
                \EndIf
            \EndFor

            \LeftComment{Connect all new input and output nodes}
            \ForAll{InputNode i in sn.inputNodes}
                \If{i.param in targetInputs}
                    \ForAll{OutputNode o in sn.outputNodes}
                        \State sn.addEdge(i, o, $weight \gets \mathcal{U}(-0.5,0.5)$)
                    \EndFor
                \EndIf
            \EndFor

            \ForAll{OutputNode o in sn.outputNodes}
                \If{o.param in targetOutputs}
                    \ForAll{InputNode i in sn.inputNodes}
                        \State sn.addEdge(i, o, $weight \gets \mathcal{U}(-0.5,0.5)$)
                    \EndFor
                \EndIf
            \EndFor
        \EndFunction
    \end{algorithmic}

    \caption{\label{alg:original_transfer} ASTL Seed Network Adaptation}
\end{algorithm}

\subsection{Network-Aware Adaptive Structure Transfer Learning}
\label{sec:nastl}
While ASTL had some preliminary success in transferring RNNs for time series modeling tasks, it only added connections between input and output nodes and ignored the internal latent structure of the network being transferred. Furthermore, it re-initialized all weights in the network, only retaining the source network's structure. In this work, we generalize ASTL to a process we call \emph{network-aware} adaptive structure transfer learning (N-ASTL), which utilizes information about the seed network to improve the transfer learning process. Our hypothesis is that when using any existing RNN as a seed network, the RNN itself already contains useful information about the form of its topology as well as weight distribution information which can be used in the transfer learning process. Thus, N-ASTL leverages knowledge of a seed network's connectivity and weight distribution to inform how it connects new input and output nodes. N-ASTL involves three strategies, detailed below.

\subsubsection{Epigenetic Weight Initialization}
\label{sec:nastl_weight_initialization}
In ASTL, new node biases and edge weights were initialized uniformly at random, $\mathcal{U}(-0.5, 0.5)$, similar to how EXAMM initializes weights in the minimal seed genome. In N-ASTL, before adapting any structure, the mean, $\mu_w$, and standard deviation, $\sigma_w$, of the seed network's weights are computed. Afterwards, when new edges are generated during the seed network adaption process, weights are initialized according to a dynamic normal (Gaussian) distribution driven by $\mu_w$ and $\sigma_w$, or $\mathcal{N}(\mu_w, \sigma_w)$. This mirrors how EXAMM performs epigenetic/Lamarckian weight/bias initialization when performing mutation and crossover operations.

\begin{algorithm}[ht]
    \begin{algorithmic}
        \Function{NASTL-Inputs}{SeedNetwork sn, Param[] targetOutputs, Param[] targetInputs}
            \State \Call{RemoveUnused}{sn, targetOutputs, targetInputs}

            \State $\mu_w \gets $ sn.getWeightMean()
            \State $\sigma_w \gets $ sn.getWeightStdDev()

            \State $\mu_o \gets $ sn.getMeanOutputs()
            \State $\sigma_o \gets $ sn.getStdDevOutputs()
        
            \LeftComment{Connect the new input nodes}
            \ForAll{InputNode i in sn.inputNodes}
                \If{i.param in targetInputs}
                    \State $nInputs \gets max(1, \mathcal{N}(\mu_o, \sigma_o))$

                    \State Node[] nodes $\gets$ sn.getEnabledHiddenNodes() $\cup$ sn.getOutputNodes()
                    \State \Call{shuffle}{nodes}

                    \For{$j \gets 1 \text{ to } nInputs$}
                        \State sn.addEdge(i, nodes[j], $weight \gets \mathcal{N}(\mu_w,\sigma_w)$)
                    \EndFor
                \EndIf
            \EndFor
        \EndFunction
    \end{algorithmic}

    \caption{\label{alg:nastl_inputs} N-ASTL Seed Network Adaptation: Output-Aware Input Connection}
\end{algorithm}

\subsubsection{Output-Aware Input Connections}
\label{sec:nastl_input_connection}
Algorithm~\ref{alg:nastl_inputs} presents the output-aware input connection procedure for N-ASTL. Similar to our dynamic weight initialization scheme, before adapting any structure of the seed network, the mean, $\mu_o$, and standard deviation, $\sigma_o$, of the number of outputs that each input and hidden node has in the network are calculated. Following this, the unused input and output nodes are then removed from the seed network, with any resulting vestigial hidden nodes and edges appropriately disabled. Following this, the new output and input nodes are added to the network. Each new input node is connected to either output nodes or enabled hidden nodes with a number of connections that is randomly selected according to a Gaussian distribution but with the restriction that at least one connection must be made, \ie, $max(1, \mathcal{N}(\mu_o,\sigma_o))$. This ensures that all input nodes are connected to the seed network in a functional way that also follows a similar distribution to the seed RNN's existing structure.

\subsubsection{Input-Aware Output Connections}
\label{sec:nastl_output_connection}
Algorithm~\ref{alg:nastl_outputs} presents the input-aware output connection procedure for N-ASTL. New output nodes are connected in a way similar to that used for the new input nodes. Before any adaptation, the mean, $\mu_i$, and standard deviation, $\sigma_i$, of the number of inputs that each output and hidden node has in the network is calculated. After removing unused input and output nodes (along with disabling vestigial edges/hidden nodes) and then adding the new input and output nodes, output nodes are potentially wired to any input node or enabled hidden node. The number of connections is then sampled from a Gaussian distribution over the number of inputs, again with the restriction that at least one connection is made, i.e., $max(1,\mathcal{N}(\mu_i,\sigma_i))$.

\begin{algorithm}[ht]
    \begin{algorithmic}
        \Function{NASTL-Outputs}{SeedNetwork sn, Param[] targetOutputs, Param[] targetInputs}
            \State \Call{RemoveUnused}{sn, targetOutputs, targetInputs}

            \State $\mu_w \gets $ sn.getWeightMean()
            \State $\sigma_w \gets $ sn.getWeightStdDev()

            \State $\mu_i \gets $ sn.getMeanInputs()
            \State $\sigma_i \gets $ sn.getStdDevInputs()
        
            \LeftComment{Connect the new output nodes}
            \ForAll{OutputNode o in sn.outputNodes}
                \If{o.param in targetOutputs}
                    \State $nOutputs \gets max(1, \mathcal{N}(\mu_i, \sigma_i))$

                    \State Node[] nodes $\gets$ sn.getEnabledHiddenNodes() $\cup$ sn.getInputNodes()
                    \State \Call{shuffle}{nodes}

                    \For{$j \gets 1 \text{ to } nOutputs$}
                        \State sn.addEdge(nodes[j], o, $weight \gets \mathcal{N}(\mu_w,\sigma_w)$)
                    \EndFor
                \EndIf
            \EndFor
        \EndFunction
    \end{algorithmic}

    \caption{\label{alg:nastl_outputs} N-ASTL Seed Network Adaptation: Input-Aware Output Connection}
\end{algorithm}

\begin{figure*}[ht]
    \centering
    \subfloat[Cessna 172 Skyhawk\label{fig:c172}]{
        \includegraphics[width=0.30\textwidth]{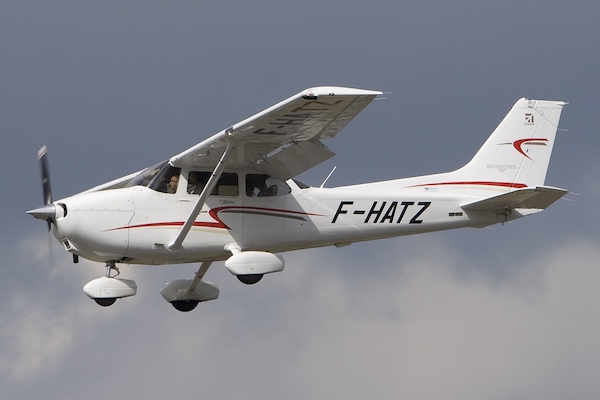}
    }\hfill
    \subfloat[Piper PA-28 Cherokee\label{fig:pa28}]{
        \includegraphics[width=0.30\textwidth]{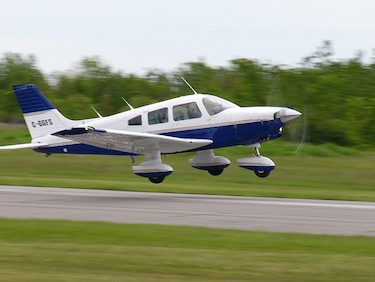}
    }\hfill
    \subfloat[Piper PA-44 Seminole\label{fig:pa44}]{
        \includegraphics[width=0.30\textwidth]{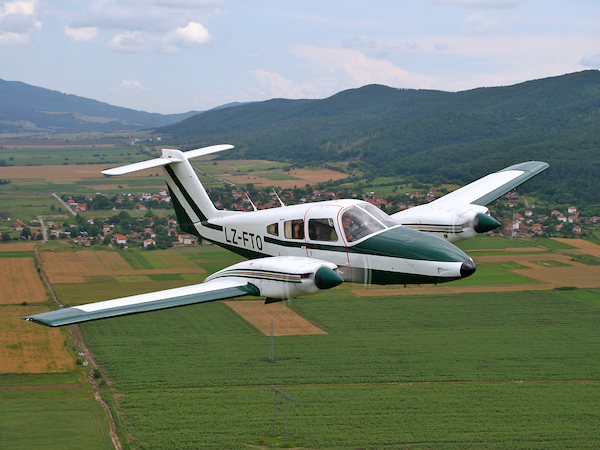}
    }
    \caption{\label{fig:airframes} The data used for transfer learning comes from three different airframes (images under creative commons licenses).}
\end{figure*}

\begin{figure*}
    \centering
    \includegraphics[width=0.98\textwidth]{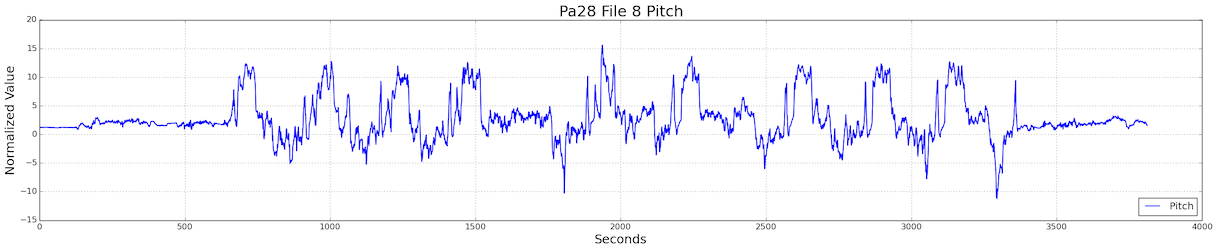}
    \\
    \includegraphics[width=0.98\textwidth]{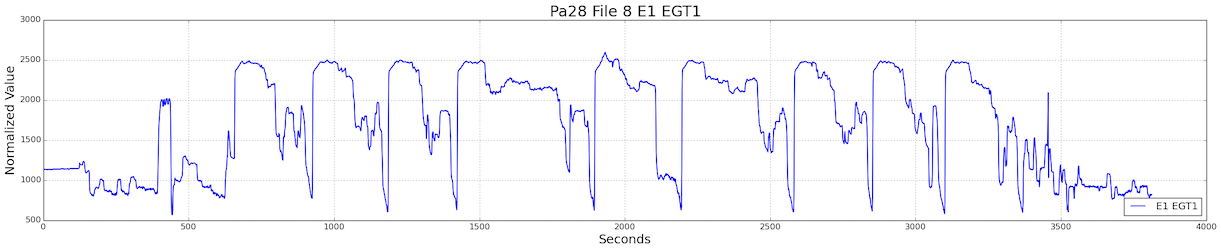}
    \caption{\label{fig:flight_examples} Example of the pitch and E1 EGT1 parameters from PA28 flight $8$ in the NGAFID dataset.}
\end{figure*}

\section{Transfer Learning Tasks}
\label{sec:data}

To compare to prior state of the art results reported for the original ASTL, we utilize the same open source aviation based dataset which has been provided as part of the EXAMM github repository\footnote{https://github.com/travisdesell/exact/tree/master/datasets/2019\_ngafid\_transfer}. The source data for this transfer learning problem consists of $36$ flights gathered from the National General Aviation Flight Information Database\footnote{http://ngafid.org}. It includes three different airframes, with $12$ flights coming from Cessna 172 Skyhawks (C172s), $12$ from Piper PA-28 Cherokees (PA28s), and the last $12$ from Piper PA-44 Seminoles (PA44s). Each of the $36$ flights came from a different aircraft. The different airframes have significant design differences (see Figure~\ref{fig:airframes}). The C172s are ``high wing'' (wings are on the top) with a single engine, the PA28s are ``low wing'' (wings are on the bottom) and have a single engine, and the PA44s are low wing with dual engines. The flight data consists of per second readings and the duration of each flight data file ranges from $1$ to $3$ hours. Each airframe shares $18$ common sensor parameters, C172 and PA44 add 7 additional sensor parameter which the PA28 does not have, C172s have 3 additional engine parameters which PA-28s and PA-44s do not have, and PA-44s add an additional $11$ parameters, mostly related to its second engine, which C172s and PA-28s do not have. All available parameters were used as inputs and Appendix~\ref{sec:dataset_details} provides a detailed tabular description of which sensors each airframe has and which were used as prediction outputs (if available). Figure~\ref{fig:flight_examples} provides an example of the data being predicted showing the pitch and E1 EGT1 values from PA28 flight $8$, illustrating the challenges involved. The data is very noisy containing sudden, non-stationary, and non-seasonal changes as well as varying correlations to the other input parameters. As a result, traditional statistical methods, \eg, those from the auto-regressive integrated moving average (ARIMA) family of models, are not well-suited to the task.
\section{Results}
\label{sec:results}

\begin{figure*}[ht]
\centering
    \subfloat[\label{fig:c172_pa28_transfer} C172 to PA28]{
        \includegraphics[width=0.475\textwidth]{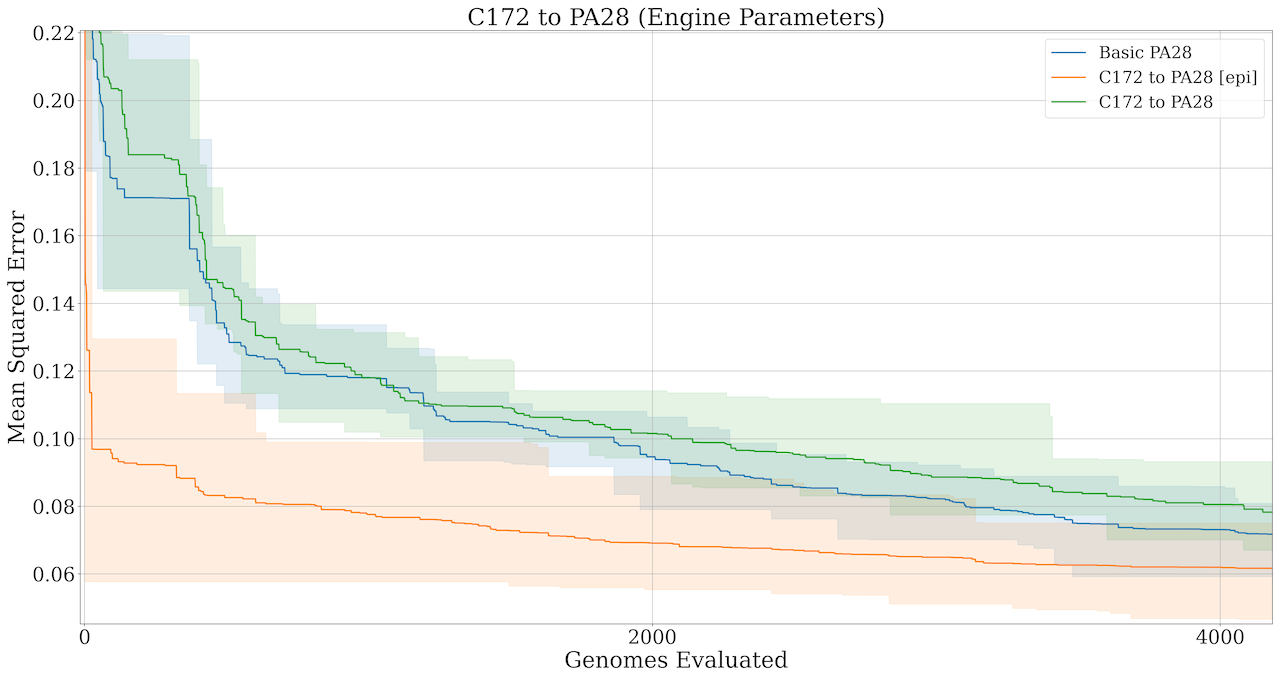}
    }\hfill
    \subfloat[\label{fig:pa44_pa28_transfer} PA44 to PA28]{
        \includegraphics[width=0.475\textwidth]{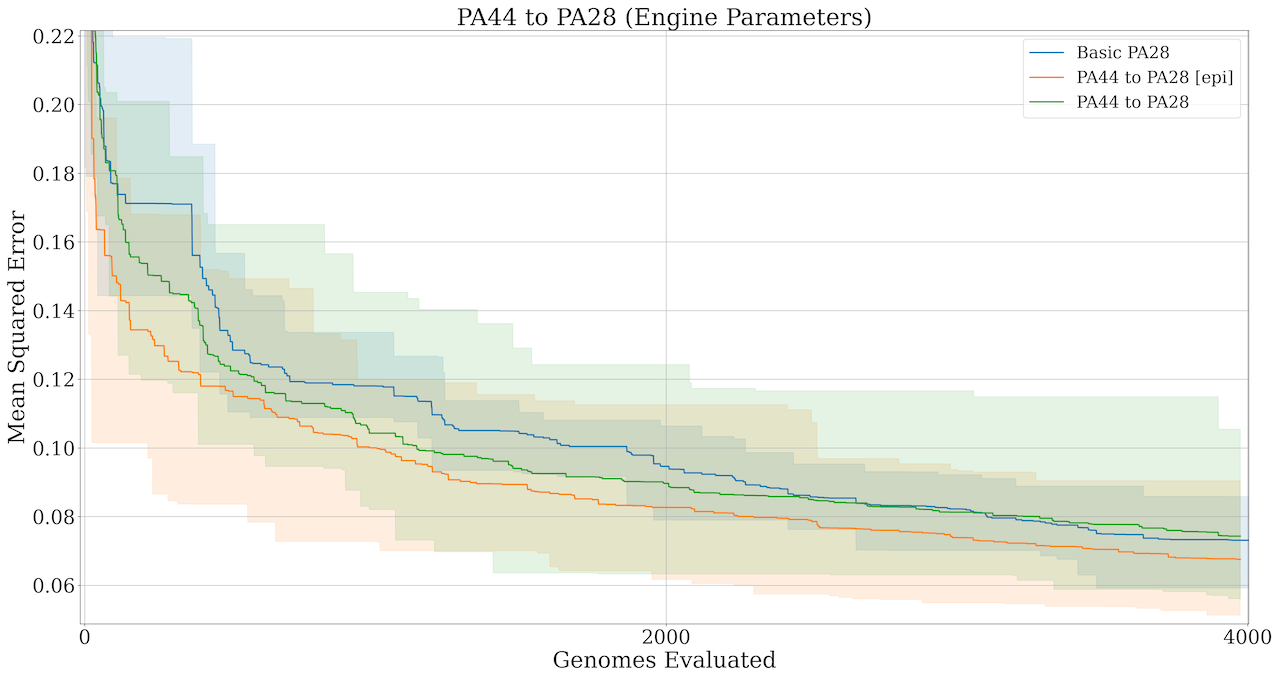}
    }

    \subfloat[\label{fig:pa28_c172_transfer} PA28 to C172]{
        \includegraphics[width=0.475\textwidth]{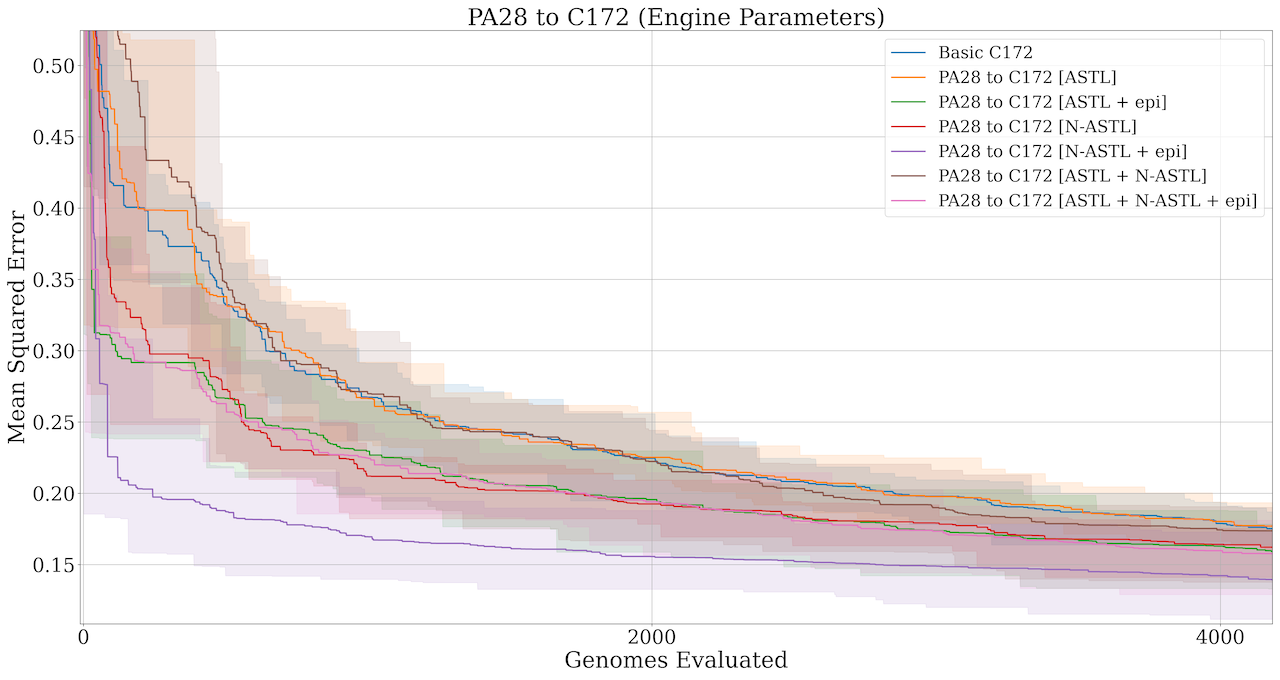}
    }\hfill
    \subfloat[\label{fig:pa44_c172_transfer} PA44 to C172]{
        \includegraphics[width=0.475\textwidth]{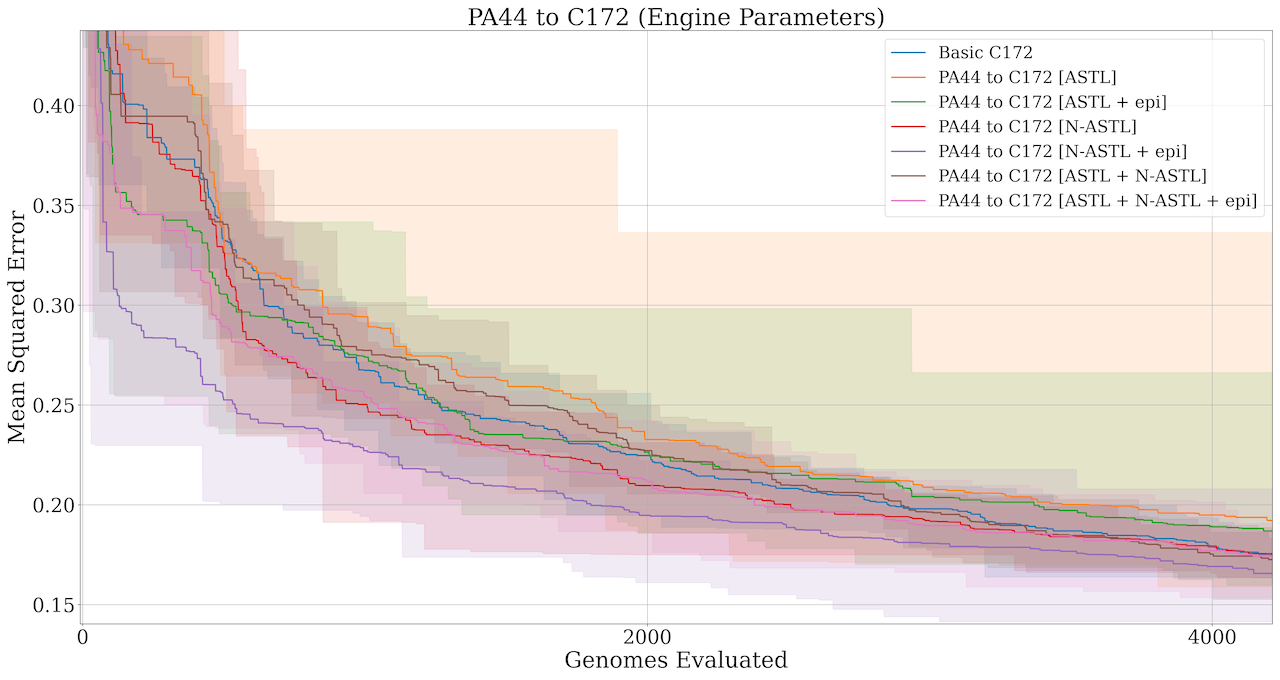}
    }

    \subfloat[\label{fig:pa28_pa44_transfer} PA28 to PA44]{
        \includegraphics[width=0.475\textwidth]{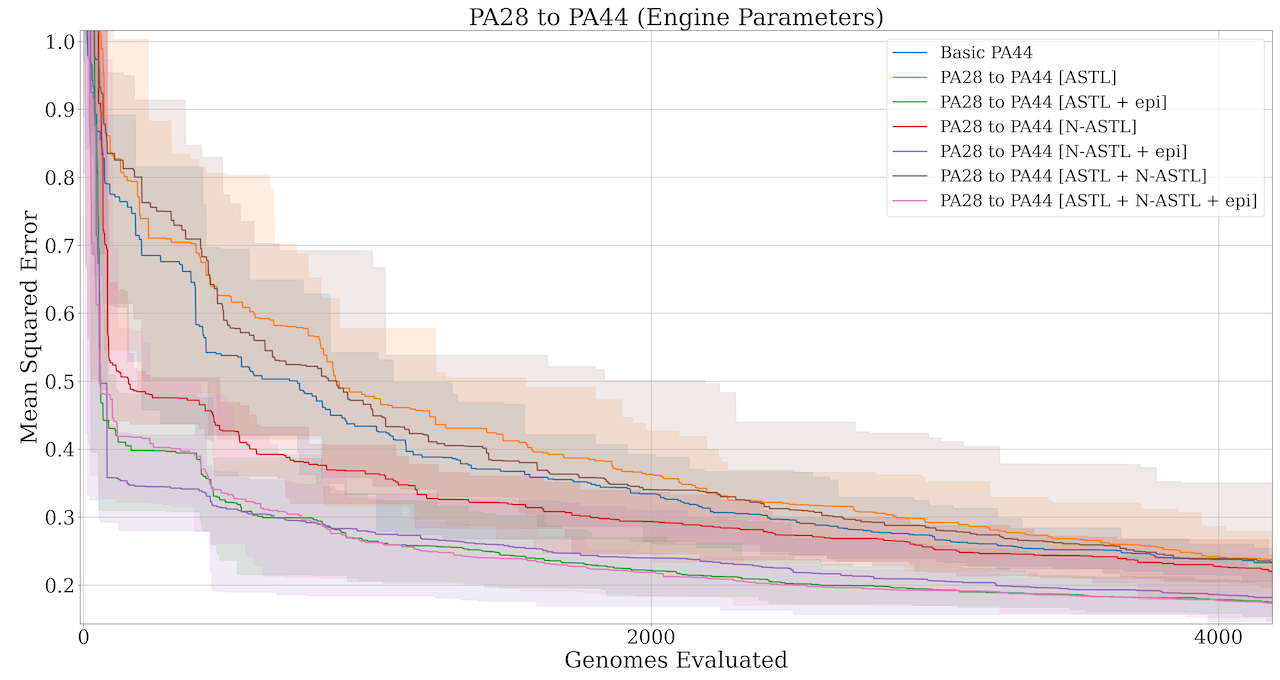}
    }\hfill
    \subfloat[\label{fig:c172_pa44_transfer} C172 to PA44]{
        \includegraphics[width=0.475\textwidth]{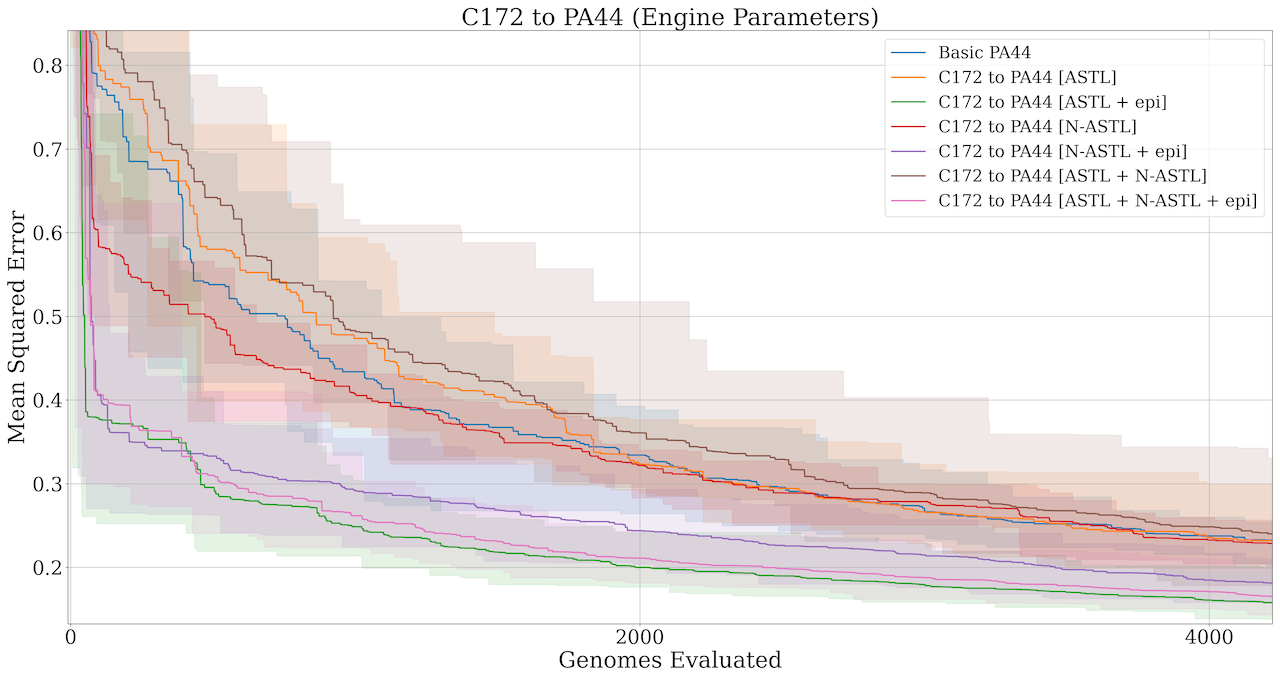}
    }
    \caption{\label{fig:from_scratch_transfer} Convergence rates (in terms of best MSE on validation data) for the EXAMM runs starting from scratch (Basic C172, PA28 or PA44) compared to starting with a seed network transferred from a different airframe when predicting engine exhaust gas temperature (EGT) values.}
\end{figure*}

\begin{figure*}[ht]
\centering
    \subfloat[\label{fig:c172_pa28_transfer_4k} C172 to PA28]{
        \includegraphics[width=0.475\textwidth]{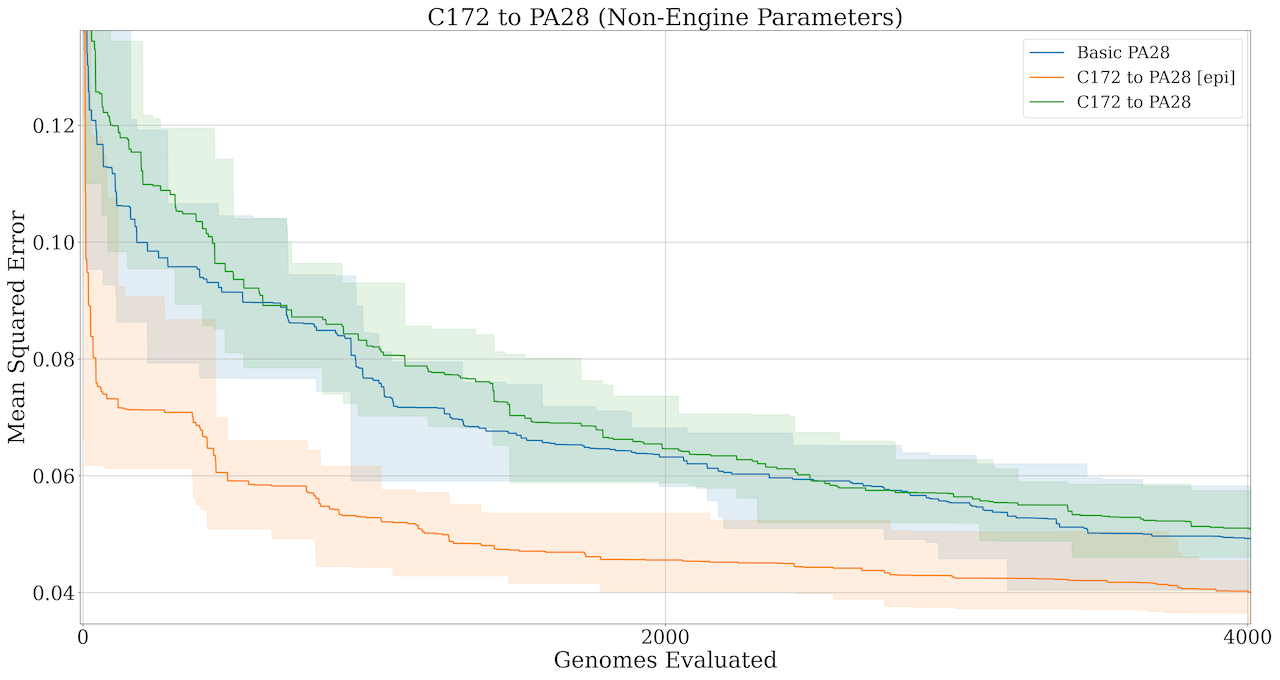}
    }\hfill
    \subfloat[\label{fig:pa44_pa28_transfer_4k} PA44 to PA28]{
        \includegraphics[width=0.475\textwidth]{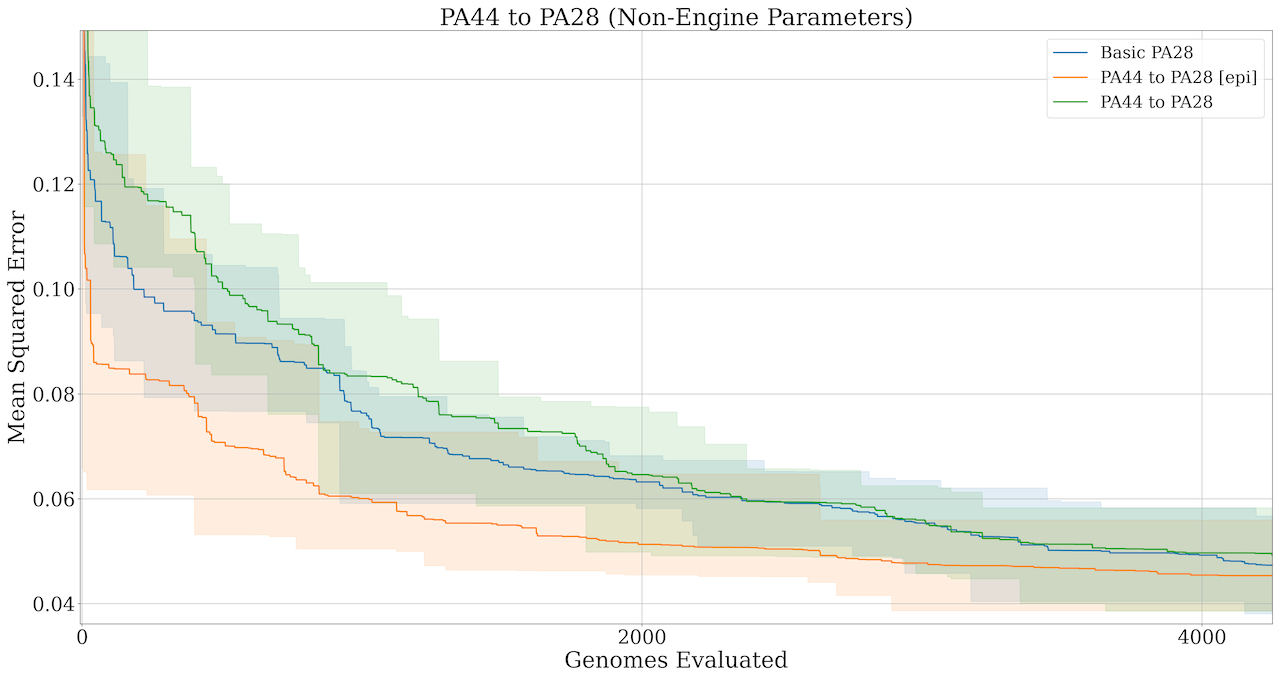}
    }

    \subfloat[\label{fig:pa28_c172_transfer_4k} PA28 to C172]{
        \includegraphics[width=0.475\textwidth]{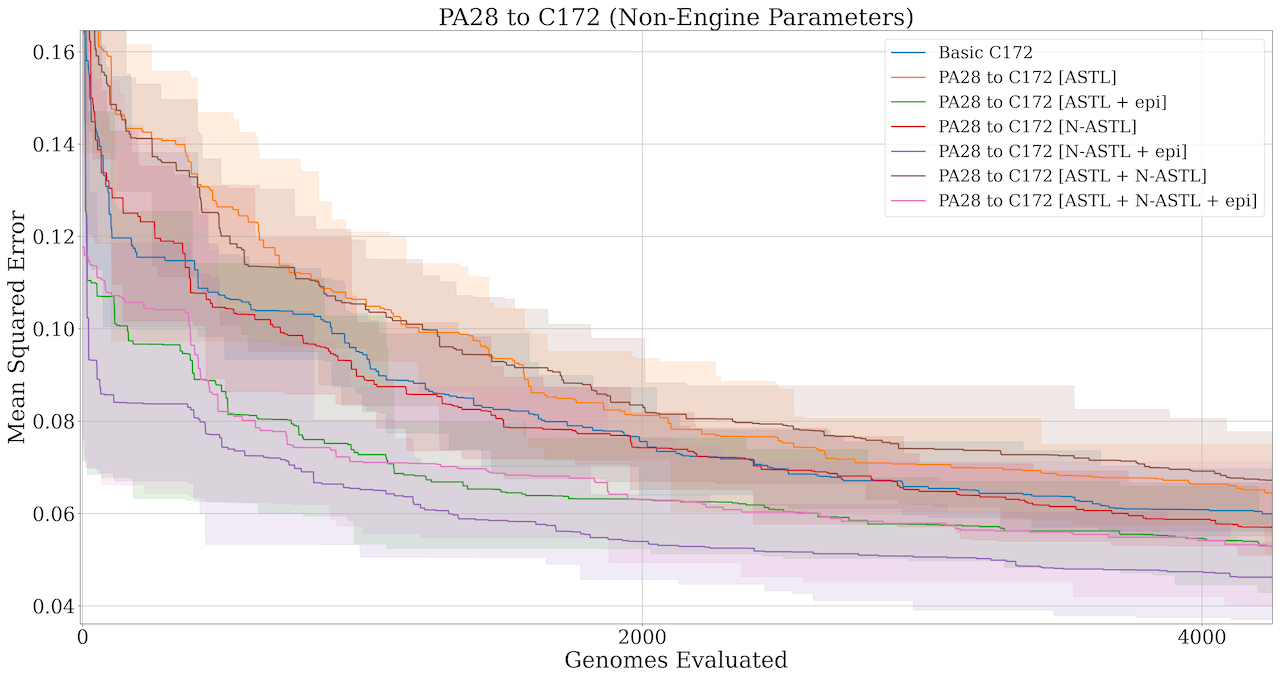}
    }\hfill
    \subfloat[\label{fig:pa44_c172_transfer_4k} PA44 to C172]{
        \includegraphics[width=0.475\textwidth]{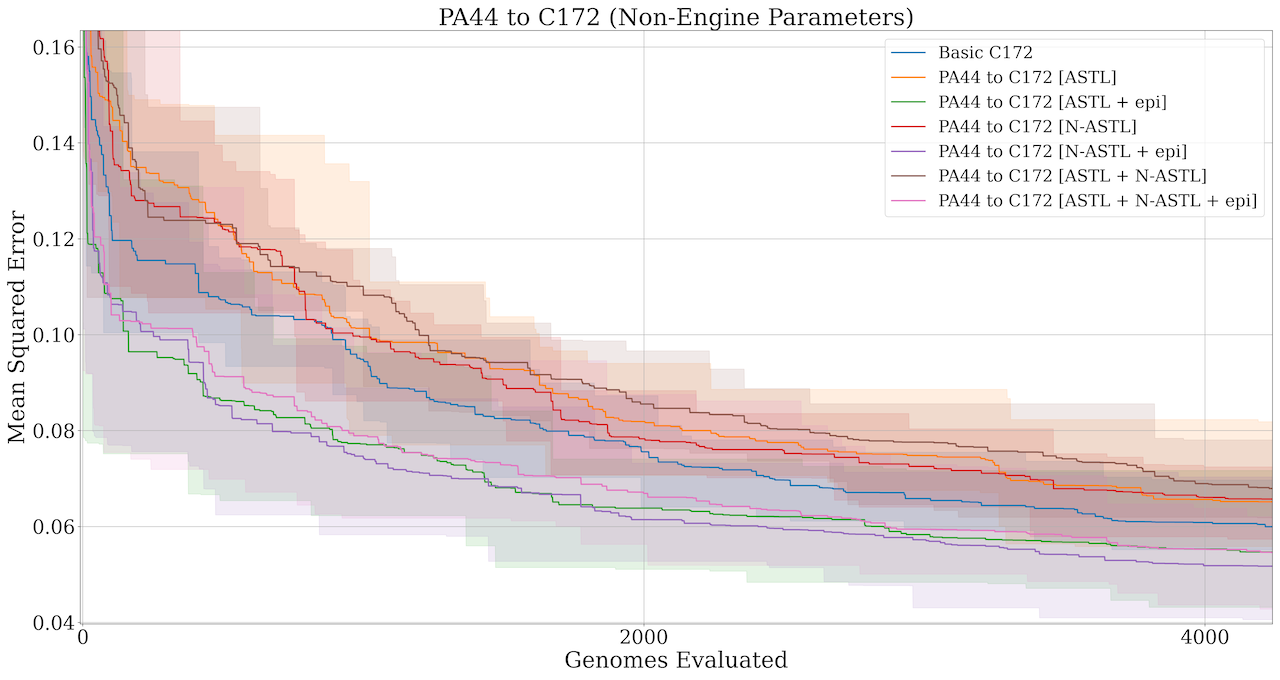}
    }

    \subfloat[\label{fig:pa28_pa44_transfer_4k} PA28 to PA44]{
        \includegraphics[width=0.475\textwidth]{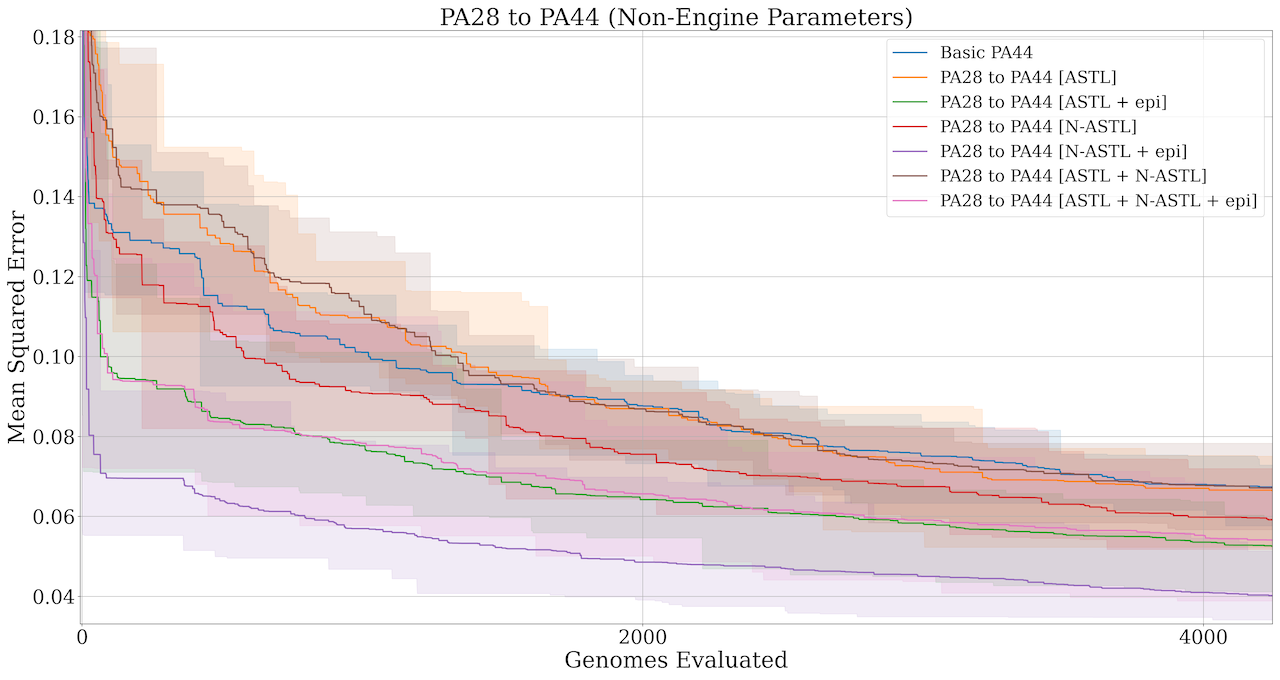}
    }\hfill
    \subfloat[\label{fig:c172_pa44_transfer_4k} C172 to PA44]{
        \includegraphics[width=0.475\textwidth]{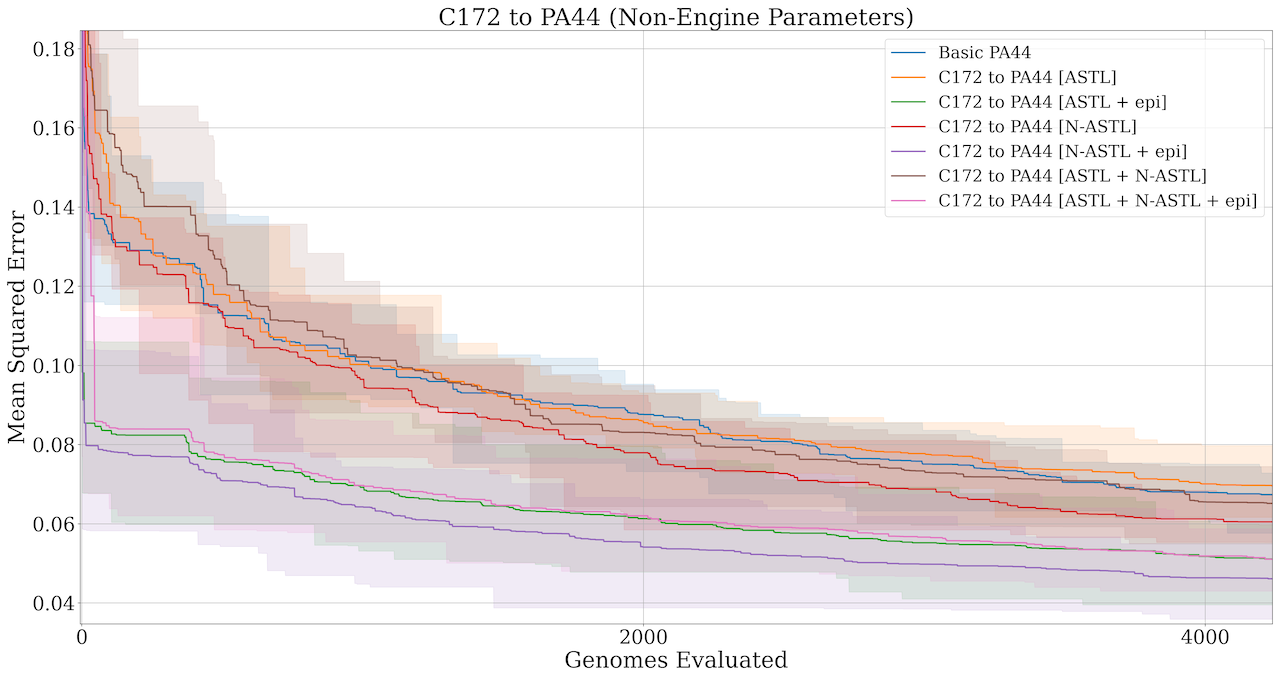}
    }
    \caption{\label{fig:transfer_4k} Convergence rates (in terms of best MSE on validation data) for the EXAMM runs starting from scratch (Basic C172, PA28 or PA44) compared to starting with a seed network transferred from a different airframe predicting the non-engine parameters (AltAGL, IAS, LatAc, NormAc, Pitch and Roll).}
\end{figure*}

\subsection{Hyperparameter Settings}
\label{sec:examm_bp_hyperparameters}

All EXAMM neuro-evolution runs utilized $4$ islands, each with a max population size of $10$. New RNNs were generated via a mutation rate of 70\%, an intra-island crossover rate of 20\%, and an inter-island crossover rate of 10\%. $10$ out of EXAMM's $11$ mutation operations were utilized (all except for \emph{split edge}), each chosen with a uniform 10\% chance. EXAMM generated new nodes by selecting from a set that included simple neurons, $\Delta$-RNN, GRU, LSTM, MGU, and UGRNN memory cells uniformly at random. Recurrent connections could span any time-skip, sampled uniformly, i.e., $\mathcal{U}(1,10)$.

All RNNs were locally trained for $4$ epochs via stochastic gradient descent (SGD) using backpropagation through time (BPTT)~\cite{werbos1990backpropagation} to compute gradients, all using the same hyperparameters. RNN weights were initialized via EXAMM's Lamarckian strategy (described in~\cite{ororbia2019examm}), which allows children RNNs to reuse parental weights, significantly reducing the number of epochs required for the neuroevolution's local RNN training steps. SGD was run with a learning rate of $\eta = 0.001$ and used Nesterov momentum with $\mu = 0.9$. For the memory cells with forget gates, the forget gate bias had a value of $1.0$ added to it (motivated by \cite{jozefowicz2015empirical}).  To prevent exploding gradients, gradient clipping~\cite{pascanu2013difficulty} was used when the norm of the gradient exceeded a threshold of $1.0$. To combat vanishing gradients, gradient boosting (the opposite of clipping) was used when the gradient norm was below $0.05$. These parameters were selected by hand based on prior experience with this dataset.

\subsection{Experimental Design}
\label{sec:experiments}

One major goal of this work is to facilitate and enable the fast development of predictive systems. In the realm of aviation, this could mean that a given organization may operate a fleet of aircraft of certain airframes and employ RNN estimators as part of their predictive systems. Instead of having to train new RNNs from scratch every time existing airframes are modified, new airframes start being utilized, or sensor systems are upgraded, they can instead adapt RNNs trained on existing airframes and transfer them over for use in these new/modified systems. This transfer process would also require less data compared to the scenario of training estimators from scratch. To mirror such a scenario, we used EXAMM to evolve and train RNNs on each of the three airframes (C172s, PA28s and PA44s) for $4,000$ generated and trained RNNs, \ie~ $4,000$ genomes were evaluated. This was repeated $10$ times for each airframe. The flight data files were split up into training and validation data, with the first $9$ (of each set of flights) used for training and the last $3$ used as validation data.

\begin{table}
    \footnotesize
    \centering
    \begin{tabular}{|l|c|c|c|c|l|}
        \hline
             & Inputs & Inputs  & Outputs & Outputs \\
        Task & Added  & Removed & Added   & Removed \\
        \hline
        PA28 to PA44  & 13 & 0  & 4 & 0 \\
        PA28 to C172  & 8  & 0  & 3 & 0 \\
        C172 to PA28  & 0  & 8  & 0 & 3 \\
        C172 to PA44  & 10 & 3  & 4 & 0 \\
        PA44 to PA28  & 0  & 13 & 0 & 7 \\
        PA44 to C172  & 3  & 10 & 0 & 7 \\
        \hline
    \end{tabular}
    \caption{\label{table:tl_tasks} Transfer Learning Tasks}
    \vspace{-5mm}
\end{table}

To appropriately evaluate our proposed N-ASTL methodology, we utilized the same prediction task defined in~\cite{elsaid2019evostar-transfer}, which was to predict the exhaust gas temperature (EGT) engine parameters for the target airframe, as well as a new task that entailed predicting non-engine parameters. RNNs predicting on PA28 would predict engine 1 (E1) EGT1, those for C172s would predict E1 EGT1-4, and RNNs predicting on PA44 data would predict both E1 EGT1-4 and engine 2 (E2) EGT1-4. The non-engine prediction parameters were Altitude Miles Above Sea Level (AltMSL), Indicated Air Speed (IAS), Lateral Acceleration (LatAc), Normal Acceleration (NormAc), Pitch and Roll. We investigated the transfer learning methods using each airframe as a source and target, resulting in six different transfer learning examples: C172 to PA28, C172 to PA44, PA28 to C172, PA28 to PA44, PA44 to C172, and PA44 to PA28. 

Table~\ref{table:tl_tasks} presents the different transfer learning tasks examined, along with how many input and output parameters were added and removed in each example. Input nodes were added or removed by the previously described strategies to utilize all available sensor inputs for the target data. Likewise, outputs were added or removed to predict all the available EGT parameters. For the non-engine parameters no outputs needed to be added or removed. For each of the $10$ repeated runs, the best genome in each set after the $4,000$ genome evaluations was selected as a seed network for the transfer learning strategies. For the $3$ airframes, the $10$ selected seed networks were then utilized to evaluate the different adaptation strategies. We examined ASTL and N-ASTL independently, as well as using them together (denoted as {\tt ASTL + N-ASTL}) for connecting the new input and output nodes. For each of these three strategies, we used either ASTL weight initialization or N-ASTL epigenetic weight initialization (denoted with {\tt +epi}). For the runs where PA28 was the target, since we only removed inputs and outputs, we only examined ASTL with and without epigenetic weight initialization, as there were no new nodes to connect.

\subsection{Structural Adaptation Evaluation}

Figures~\ref{fig:from_scratch_transfer} and~\ref{fig:transfer_4k} compare the different seed network adaptation strategies and their combinations using each of the airframes (C172, PA28 and PA44) as a source to be transferred to each of other airframes (the targets) evolved to predict the engine parameter values and non-engine parameter values, respectively. 

For the experiments using the PA28 data as a target, since no input or output nodes were added, transfer learning was tested with and without the N-ASTL epigenetic weight initialization. In all four cases, epigenetic weight initialization had strong error reduction and learned the task much quicker. Additionally, improved results were seen transferring from the PA44 RNNs as a source, whereas the prior ASTL results had failed.

For the experiments using the C172 data as a target, large improvements were again seen with epigenetic weight initialization. In all cases, {\tt N-ASTL+epi} learned the quickest and resulted in the lowest error, with {\tt N-ASTL+ASTL+epi} providing the next best results followed by {\tt ASTL+epi}. Even without epigenetic weights, N-ASTL outperformed the strategies involving ASTL connectivity.

For the experiments using the PA44 data as a target, N-ASTL strongly shows its significance. Prior results with ASTL were unable to improve over utilizing EXAMM from scratch on the PA44 data, while in this case, even without epigenetic weights, N-ASTL is still able to improve over EXAMM from scratch.  Additionally, the use of epigenetic weights shows a very large improvement, with all three versions showing significant improvement over N-ASTL. In three of the four cases, again {\tt N-ASTL+epi} learned the fastest and found the most accurate results, except in the case of C172 to PA44 on the engine parameter predictions, where {\tt ASTL+epi} and {\tt N-ASTL+ASL+epi} performed comparatively, though slightly, better.

\subsection{Conclusions}
Overall, we find these results strongly positive as they show that the N-ASTL strategies yield significantly better performance across all experiments. Furthermore, in almost every case, the N-ASTL strategies created better performing RNNs after $2,000$ genomes compared to the best found after $4,000$ genomes when evolving RNNs ``from scratch''. If we look at the curvature of these plots, the RNNs that evolved from scratch and the ASTL and non-epigenetic weight tests were converging to significantly worse performance than the N-ASTL runs. This suggests that transferring RNNs trained on other data and seeding them with genomes in a network-aware manner produces RNN predictors that generalize far better.  

Epigenetic weight initialization significantly improved the results in all cases, showing that utilizing network aware weight distributions for weight initialization is highly important for the transfer learning process. Additionally, in $6$ of the $8$ cases that involved adding/removing inputs and outputs (\ie, those transferring to C172 or PA44), N-ASTL without epigenetic weight initialization outperformed RNNs that were evolved/trained from scratch, suggesting that utilizing network-aware topology information is also important to the process.  This is further backed by the fact that in $6$ of those $8$ cases the {\tt N-ASTL+epi} runs provided the best results. For the two cases they where did not, PA28 to PA44 and C172 to PA44, which were when additional outputs were added, results suggest that when additional outputs are added, making sure they are connected to all inputs is important. On the other hand, making sure new inputs are connected to all existing outputs is not as important.

\section{Discussion}
\label{sec:discussion}

This work investigates the use of a novel network-aware adaptive structure transfer learning strategy (N-ASTL) to further speed transfer learning of deep RNNs. N-ASTL utilizes statistical information about the source RNN's topology and weight distributions to inform how it should be adapted to new data sets which have different input and output parameters, necessitating the use of a different neural architecture. These strategies were evaluated using the challenging real world problem of performing transfer learning of RNNs trained to predict aviation engine parameters between three different airframes with different designs and engines. 

N-ASTL provided significant performance improvements over prior state of the art, which did not take into account network topology or weight distributions. Further, N-ASTL was shown to be able to successfully perform transfer learning on tasks where transfer learning was not previously possible. Interestingly enough, this work shows that in many cases the transfer learning strategies are able to evolve RNNs that outperform ones which started from scratch but were evolved and trained on the target dataset for twice as long. In many cases, the curvature of those plots suggest they would never reach the performance of the RNNs seeded by a transferred network. This suggests that the transfer learning strategy is able to evolve more robust and generalizable RNNs, as performance of the non-transferred RNNs levels off at a significantly lower accuracy than the transfer learning evolved RNNs. These results are significant and showcase the use of transfer learning as a means to enhance predictive systems in aviation, with applications to other domains involving time series data of different input and output dimensionalities. 

This study also opens up a number of directions for future work. While N-ASTL only seeds EXAMM with a single adapted network structure, the manner in which it connects new inputs and outputs is stochastic. This provides the potential to generate multiple seed network candidates to provide more initial variety to EXAMM's island populations which could lead to improved reliability and robustness when searching for optimal RNN architectures. Additionally, there appears to be a difference between adding inputs to adding outputs. While the tests which only added inputs but not outputs had N-ASTL with epigenetic weights achieving the best results, the tests which added outputs (those transferring to PA44) showed better performance when also adding in ASTL connectivity. It is worth examining if using N-ASTL plus a modified version of ASTL which only fully connects new outputs to inputs may prove better in these cases.  Further, N-ASTL has shown that connecting new inputs and outputs to nodes in the hidden layers is important. It is worth further study to see if simply connecting new inputs and outputs to \emph{all} hidden nodes provides any benefit.

Lastly, while this work has focused on the challenging problem of time series prediction with RNNs, future work will involve applying N-ASTL to RNN classification tasks, such as natural language processing, to see if transferring between different language dictionaries or word and character embeddings provides similar improvements, as well as to convolutional neural networks, allowing transfer learning between images and output spaces of different shapes and sizes.


\section{Acknowledgements}
This material is based upon work supported by the U.S. Department of Energy, Office of Science, Office of Advanced Combustion Systems under Award Number \#FE0031547 and by the Federal Aviation Administration and MITRE Corporation under the National General Aviation Flight Information Database (NGAFID) award.

\newpage
\bibliographystyle{unsrt}
\bibliography{references} 

\newpage

\onecolumn
\begin{appendices}
\section{Aviation Dataset Parameters}
\label{sec:dataset_details}

This paper utilizes 12 flight logs each from aircraft of three different airframes: Cessna 172 Skyhawk (C172s), Piper-Archer 28 Cherokees (PA-28s) and Piper-Archer 44 Seminoles (PA-44s).  These aircraft share $18$ common sensor parameters, and then each has varying sensor parameters for their engine(s). The data files used in this work are freely available as comma separated value (CSV) format as part of the EXAMM github repository\footnote{https://github.com/travisdesell/exact/tree/master/datasets/2019\_ngafid\_transfer}. The following table presents which sensors are present for which airframe and which were used as prediction outputs (if available), all available parameters were used as inputs.:

\vspace{5mm}
\begin{center}

\begin{minipage}{1\textwidth}
\scriptsize
\centering
\begin{tabular}{|l|c|c|c|c|c|}
\hline
               & Cessna 172 & Piper-Archer 28 & Piper-Archer 44 & \multicolumn{2}{|c|}{Potential Output} \\
Parameter Name & Skyhawk    & Cherokee        & Seminole        & (Engine) & (Non-Engine) \\
\hline
Altitude Above Ground Level (AltAGL)        & x & x & x & & \\
Barometric Altitude (AltB)                  & x & x & x & & \\
GPS Altitude (AltGPS)                       & x & x & x & & \\
Altitude Miles Above Sea Level (AltMSL)     & x & x & x & & x \\
Fuel Quantity Left (FQtyL)                  & x & x & x & & \\
Fuel Quantity Right (FQtyR)                 & x & x & x & & \\
Ground Speed (GndSpd)                       & x & x & x & & \\
Indicated Air Speed (IAS)                   & x & x & x & & x \\
Lateral Acceleration (LatAc)                & x & x & x & & x \\
Normal Acceleration (NormAc)                & x & x & x & & x \\
Outside Air Temperature (OAT)               & x & x & x & & \\
Pitch                                       & x & x & x & & x \\
Roll                                        & x & x & x & & x \\
True Airspeed (TAS)                         & x & x & x & & \\
Vertical Speed (VSpd)                       & x & x & x & & \\
Vertical Speed Gs (VSpdG)                   & x & x & x & & \\
Wind Direction (WndDir)                     & x & x & x & & \\
Wind Speed (WndSpd)                         & x & x & x & & \\
\hline
Absolute Barometric Pressure (BaroA)                & x &   & x & & \\
Engine 1 Cylinder Head Temperature 1 (E1 CHT1)      & x &   & x & & \\
Engine 1 Cylinder Head Temperature 2 (E1 CHT2)      & x &   &   & & \\
Engine 1 Cylinder Head Temperature 3 (E1 CHT3)      & x &   &   & & \\
Engine 1 Cylinder Head Temperature 4 (E1 CHT4)      & x &   &   & & \\
Engine 1 Exhaust Gas Temperature 1 (E1 EGT1)        & x & x & x & x & \\
Engine 1 Exhaust Gas Temperature 2 (E1 EGT2)        & x &   & x & x & \\
Engine 1 Exhaust Gas Temperature 3 (E1 EGT3)        & x &   & x & x & \\
Engine 1 Exhaust Gas Temperature 4 (E1 EGT4)        & x &   & x & x & \\
Engine 1 Fuel Flow (E1 FFlow)                       & x & x & x & & \\
Engine 1 Oil Pressure (E1 OilP                      & x & x & x & & \\
Engine 1 Oil Temperature (E1 OilT)                  & x & x & x & & \\
Engine 1 Rotations Per minute (E1 RPM)              & x & x & x & & \\
Engine 1 Manifold Absolute Pressure (E1 MAP)        &  &  & x & & \\
Engine 2 Cylinder Head Temperature 1 (E2 CHT1)      &  &  & x & & \\
Engine 2 Exhaust Gas Temperature 1 (E2 EGT1)        &  &  & x & x & \\
Engine 2 Exhaust Gas Temperature 2 (E2 EGT2)        &  &  & x & x & \\
Engine 2 Exhaust Gas Temperature 3 (E2 EGT3)        &  &  & x & x & \\
Engine 2 Exhaust Gas Temperature 4 (E2 EGT4)        &  &  & x & x & \\
Engine 2 Fuel Flow (E2 FFlow)                       &  &  & x & & \\
Engine 2 Oil Pressure (E2 OilP)                     &  &  & x & & \\
Engine 2 Oil Temperature (E2 OilT)                  &  &  & x & & \\
Engine 2 Rotations Per minute (E2 RPM)              &  &  & x & & \\
Engine 2 Manifold Absolute Pressure (E2 MAP)        &  &  & x & & \\
\hline
\end{tabular}
\end{minipage}
\end{center}

\end{appendices}

\end{document}